\documentclass{article}

\usepackage{arxiv}

\usepackage[utf8]{inputenc} 
\usepackage[T1]{fontenc}    
\usepackage{hyperref}       
\usepackage{url}            
\usepackage{booktabs}       
\usepackage{amsfonts}       
\usepackage{nicefrac}       
\usepackage{microtype}      
\usepackage{lipsum}		
\usepackage{graphicx}
\usepackage{natbib}
\usepackage{doi}
\usepackage{pdflscape}
\usepackage{longtable}
\usepackage{geometry}
\usepackage{amsmath}
\usepackage{amssymb}
\usepackage{latexsym}

\usepackage{url}
\usepackage{xcolor}
\usepackage{booktabs}
\usepackage{authblk}
\usepackage{adjustbox}
\usepackage{longtable}

\hypersetup{
    colorlinks=true,
    linkcolor=black,
    citecolor=black,
    filecolor=black,
    urlcolor=black,
}

\graphicspath{{./high-res-img/}}

\definecolor{newcolor}{rgb}{.8,.349,.1}

\title{Dermatological Diagnosis Explainability Benchmark for Convolutional Neural Networks}%
\author[1,2]{Raluca Jalaboi}
\author[1,3,4]{Ole Winther}
\author[2]{Alfiia Galimzianova}

\affil[1]{\footnotesize Department of Applied Mathematics and Computer Science at the Technical University of Denmark, Richard Petersens Plads, Building 324, DK-2800 Kongens Lyngby, Denmark}
\affil[2]{\footnotesize Medable A/S, Havnegade 25, 3., DK-1058 Copenhagen C, Denmark}
\affil[3]{\footnotesize Bioinformatics Centre, Department of Biology, University of Copenhagen, Copenhagen, Denmark}
\affil[4]{\footnotesize Center for Genomic Medicine, Rigshospitalet, Copenhagen University Hospital, Copenhagen, Denmark}
\date{}

\begin{document}
\maketitle

\begin{abstract}
In recent years, large strides have been taken in developing machine learning methods for various dermatological applications, supported in part by the widespread success of deep learning.
To date, diagnosing diseases from images is one of the most explored applications of deep learning within dermatology. 
Convolutional neural networks~(ConvNets) are the most commonly used deep learning method in medical imaging due to their training efficiency and accuracy, although they are often described as black boxes because of their limited explainability.
One popular way to obtain insight into a ConvNet's decision mechanism is gradient class activation maps (Grad-CAM).
A quantitative evaluation of the Grad-CAM explainability has been recently made possible by the release of DermXDB, a skin disease diagnosis explainability dataset which enables benchmarking the explainability performance of ConvNet architectures. 
In this paper, we perform a literature review to identify the most common ConvNet architectures used for this task, and compare their Grad-CAM explainability performance with the explanation maps provided by DermXDB.
We identified 11 architectures: DenseNet121, EfficientNet-B0, InceptionV3, InceptionResNetV2, MobileNet, MobileNetV2, NASNetMobile, ResNet50, ResNet50V2, VGG16, and Xception.
We pre-trained all architectures on an clinical skin disease dataset, and then fine-tuned them on a subset of DermXDB.
Validation results on the DermXDB holdout subset show an explainability F1 score of between 0.35-0.46, with Xception the highest explainability performance, while InceptionResNetV2, ResNet50, and VGG16 displaying the lowest.
NASNetMobile reports the highest characteristic-level explainability sensitivity, despite it's mediocre diagnosis performance.
These results highlight the importance of choosing the right architecture for the desired application and target market, underline need for additional explainability datasets, and further confirm the need for explainability benchmarking that relies on quantitative analyses rather than qualitative assessments.

\end{abstract}

\keywords{deep learning, dermatologys, explainability, benchmark, review}


\section{Introduction}
With an expected shortage of approximately ten million healthcare professionals by 2030~\citep{world2016working}, the world is facing a massive healthcare crisis.
Automation has been proposed as a solution to the scarcity of medical professionals, with the Food and Drugs Administration in the United States approving medical devices based on artificial intelligence for marketing to the public~\citep{us2018fda}. 

This development is due in part to the advancement in machine learning using unstructured data. 
Ever since~\cite{krizhevsky2017imagenet} won the ImageNet Large Scale Visual Recognition
Challenge~\citep{ILSVRC15} using a convolutional neural network (ConvNet), ConvNets have been at the forefront of machine learning based automation.
Employed primarily in healthcare for imaging applications, ConvNets have been used for disease diagnosis~\citep{gao2019convolutional}, cell counting~\citep{falk2019u}, disease severity assessment~\citep{gulshan2016development}, disease progression estimation~\citep{kijowski2020deep}, lesion or anatomical region segmentation~\citep{hesamian2019deep, ramesh2021review}, etc.
\cite{esteva2017dermatologist} were the first to demonstrate that ConvNets can achieve expert-level performance in dermatological diagnosis using dermoscopy images.
Since then, dermatology has embraced ConvNets as a solution to various diagnosis and segmentation tasks~\citep{esteva2017dermatologist, zhang2019attention, jinnai2020development, haenssle2020man, roy2022does}.

Despite these considerable advancements in medical imaging, there has not yet been a widespread adoption of machine learning based automation in the clinical workflow.
One of the main hurdles that detract from adoption is the lack of ConvNet explainability~\citep{kelly2019key}, this issue being enhanced by the recently implemented legislation aimed at ensuring that automated methods can offer an explanation into their decision mechanisms~\citep{goodman2017european}.
Different post-hoc explainability methods have been proposed as a way to explain a ConvNet's decisions~\citep{bai2021explainable, selvaraju2017grad, lundberg2017unified, ribeiro2016should}.
Gradient class activation maps (Grad-CAM) is currently the most commonly used explainability method within medical imaging, due to its intrinsic ease of interpretation and its low computational requirements.
However, validating the resulting explanations is an expensive, time consuming process that requires domain expert intervention, and thus most explainability validations are performed as small, qualitative analyses.
With the release of DermXDB~\citep{jalaboi2022dermx}, it became possible to quantitatively analyse the explainability of ConvNets trained for diagnosing six skin conditions: acne, psoriasis, seborrheic dermatitis, viral warts, and vitiligo.

The purpose of this benchmark is to provide the means to quantitatively compare the explainability of the state-of-the-art approaches to dermatological diagnosis using photographic imaging.
Our contributions are twofold: 
\begin{enumerate}
    \item We perform a comprehensive systematic review to reveal the usage of the ConvNets for the task of dermatological diagnosis using photographic images,
    \item We benchmark the identified ConvNets for diagnostic and explainability performance and compare them with eight expert dermatologists.
\end{enumerate}


\section{Background}
\subsection{Machine learning methods in dermatological diagnosis}
After the renewed interest in artificial intelligence and machine learning that started in 2012, practitioners from both academia and the industry began investigating automated methods for dermatological applications~\citep{thomsen2020systematic, jeong2022deep}.
Until 2017, the vast majority of articles applying machine learning methods on dermatological problems were using classical models such as support vector machines~\citep{liu2012distribution, sabouri2014cascade}, and linear or logistic regression~\citep{kaur2015real, kefel2016adaptable}. 
These models were trained using hand-crafted features or features extracted using classical computer vision methods such as gray-level co-occurrence matrices~\citep{shimizu2014four}, Sobel and Hessian filters~\citep{arroyo2014detection}, or HOS texture extraction~\citep{shrivastava2016computer}.
However, the main drawback of classical computer vision approaches is that hand-crafting features is an expensive, time-consuming process, while their automated extraction is too sensitive to the environmental factors of the image acquisition~(e.g. lighting, zoom).

\cite{esteva2017dermatologist} were the first to propose a ConvNet for diagnosing skin conditions from dermoscopy images.
Their ConvNet reached expert-level performance without requiring any hand-crafted features or classical computer vision models, thus paving the way towards the current popularity of ConvNets in dermatological applications.

One key component to the rise of ConvNets was the introduction of large scale dermatological datasets.
The International Skin Imaging Collaboration (ISIC) challenge dataset~\citep{codella2018skin} is one of the best known open access dermoscopy datasets, containing  25,331 images distributed over nine diagnostic categories.
Large clinical image datasets are also available for research purposes, such as SD-260~\citep{sun2016benchmark} which consists of 20,600 clinical images of 260 different skin diseases, and DermNetNZ~\citep{dermnetnz2021} which contains more than 25,000 clinical images.

Aided by the release of increasingly more performant architectures, their publicly available pre-trained weights on the ImageNet~\citep{deng2009imagenet} dataset, and the recently published public dermatological datasets, the vast majority of research contributions in machine learning applications for dermatology rely on ConvNet architectures. 
ConvNets have been extensively used in lesion diagnosis~\citep{tschandl2017pretrained, han2018deep, reshma2022deep} and lesion segmentation~\citep{yuan2017automatic, wu2022fat, baig2020deep} on different modalities relevant for the domain.
Attempts at explaining the decisions taken by ConvNets were made by several groups~\citep{tschandl2020human, tanaka2021classification}, but no quantitative analysis was performed. 

\subsection{Explainability in convolutional neural networks}
ConvNets have, from their very beginning, been notoriously difficult to interpret and explain.
Interpretability is generally considered the ability to understand the internal structure and properties of a ConvNet architecture, while explainability is defined as a ConvNet's capacity to offer plausible arguments in favour of its decision~\citep{roscher2020explainable}.
Within healthcare, explainability is especially important due to its intrinsic ability to interact with domain experts in a common vocabulary~\citep{kelly2019key}.
Although some architecture or domain-specific explainability methods exist, most medical imaging research articles employ attribution-based methods due to their ease of use and open source access~\citep{singh2020explainable, bai2021explainable}.

There are two main ways of implementing attribution-based methods: through perturbation and by using the ConvNet's gradients.
Perturbation-based methods, such as Shapley values~\citep{lipovetsky2001analysis}, LIME~\citep{ribeiro2016should}, or SharpLIME~\citep{graziani2021sharpening}, rely on modifying the original image and then evaluating the changes in the ConvNet's prediction.
For example, LIME uses a superpixel algorithm to split the image into sections, and randomly selects a subset of superpixels to occlude.
The target ConvNet then performs an inference step on the perturbed image.
This procedure is run multiple times to identify the superpixels that lead to the most drastic change in the ConvNet's prediction.
SharpLIME uses hand-crafted segmentations to split the image into relevant sections, and then proceeds with the perturbation process defined in LIME.
The main drawback of perturbation based methods is the need to run the prediction algorithm multiple times, which leads to high computational costs and long running times.

Gradient-based methods, such as saliency maps~\citep{simonyan2015very}, guided backpropagation~\citep{springenberg2014striving}, gradient class-activation maps (Grad-CAM)~\citep{selvaraju2017grad}, or layer-wise relevance propagation~\citep{bach2015pixel}, use a ConvNet's backpropagation step to identify the areas in an image that contribute the most to the prediction.
In general, gradient-based methods compute the gradient of a given input in relation to the prediction, and apply different post-processing methods to the output.
In the case of Grad-CAM, image features are extracted by forward propagating the image until the last convolutional layer.
Then, the gradient is set to 0 for all classes except the target class, and the signal is backpropagated to the last convolutional layer. 
The extracted image features that directly contribute to the backpropagated signal constitute the Grad-CAM for the given class. 
Since the analysis can be performed at the same time as the inference itself and only requires one iteration, Grad-CAM is often used in research and industrial applications~\citep{pereira2018automatic, young2019deep, tschandl2020human, hepp2021uncertainty, jalaboi2023explainable}.
Due to its popularity, in this paper we will use Grad-CAM to benchmark the explainability of commonly used ConvNet architectures.

\section{Material and methods}

\subsection{Literature review}

\begin{table}[t!]
    \centering
    \caption{Search query used on PubMed to identify the list of relevant articles. We searched for articles focused on dermatology, using deep learning methods, written in English. The query was last performed on the 20th of February 2023.}
    \begin{tabular}{lclcl}
        \toprule
        Search term & & Search term & & Search term \\
        \midrule
        (((dermatology[MeSH Terms]) OR & AND & ((neural network[MeSH Terms]) OR & AND & (English[Language]) \\
        (skin disease[MeSH Terms]) OR & & (machine learning[MeSH Terms]) OR & & \\
        (skin lesion[MeSH Terms])) & & (artificial intelligence[MeSH Terms]) OR & & \\
        & & (deep learning) OR & & \\
        & & (deep neural network) OR & & \\
        & & (convolutional neural network)) & & \\ 
         \bottomrule
    \end{tabular}
    \label{table:literature_review_query}
\end{table}

We performed a systematic literature review on PubMed, following the methodology introduced by~\cite{thomsen2020systematic}. 
The query, described in Table~\ref{table:literature_review_query}, focused on dermatological applications of deep learning.
A total of 3,650 articles were retrieved. 
We excluded articles that focused on domains other than dermatology, articles that did not include an original contribution in disease classification, articles using modalities other than photographic images, articles using methods other than ConvNets, and articles using proprietary ConvNets.

\subsection{Explainability benchmark}
\subsubsection{Explainability dataset}
For explainability benchmarking, we use DermXDB, a skin disease diagnosis explainability dataset published by~\cite{jalaboi2022dermx}.
The dataset consists of 524 images sourced from DermNetNZ~\citep{dermnetnz2021} and SD-260~\citep{sun2016benchmark}, and labeled with diagnoses and explanations in the form of visual skin lesion characteristics by eight board-certified dermatologists.
To match the Grad-CAM output, we focus on the characteristic localization task.



\subsubsection{Diagnosis evaluation}
For establishing the expert-level diagnosis performance, we compare each dermatologist with the reference standard diagnosis.
We follow the same approach for benchmarking the diagnosis performance of the ConvNets.
We evaluate the performance using the categorical F1 score, sensitivity, and specificity, defined as:

\begin{equation}
    \text{F1 score} = \frac{2TP}{2TP + FP + FN},
\end{equation}

\begin{equation}
    \text{Sensitivity} = \frac{TP}{TP + FN},
\end{equation}

\begin{equation}
    \text{Specificity} = \frac{TN}{TN + FP},
\end{equation}
where the true positives $TP$ represent correctly classified samples, the false positives $FP$ represent samples incorrectly classified as part of the target class, the false negatives $FN$ represent samples of the target class incorrectly classified as being part of a different class, and the true negatives $TN$ represent samples correctly identified as not being part of the target class.

\subsubsection{Explainability evaluation}
\begin{figure}[tb]
    \centering
    \includegraphics[width=0.55\textwidth]{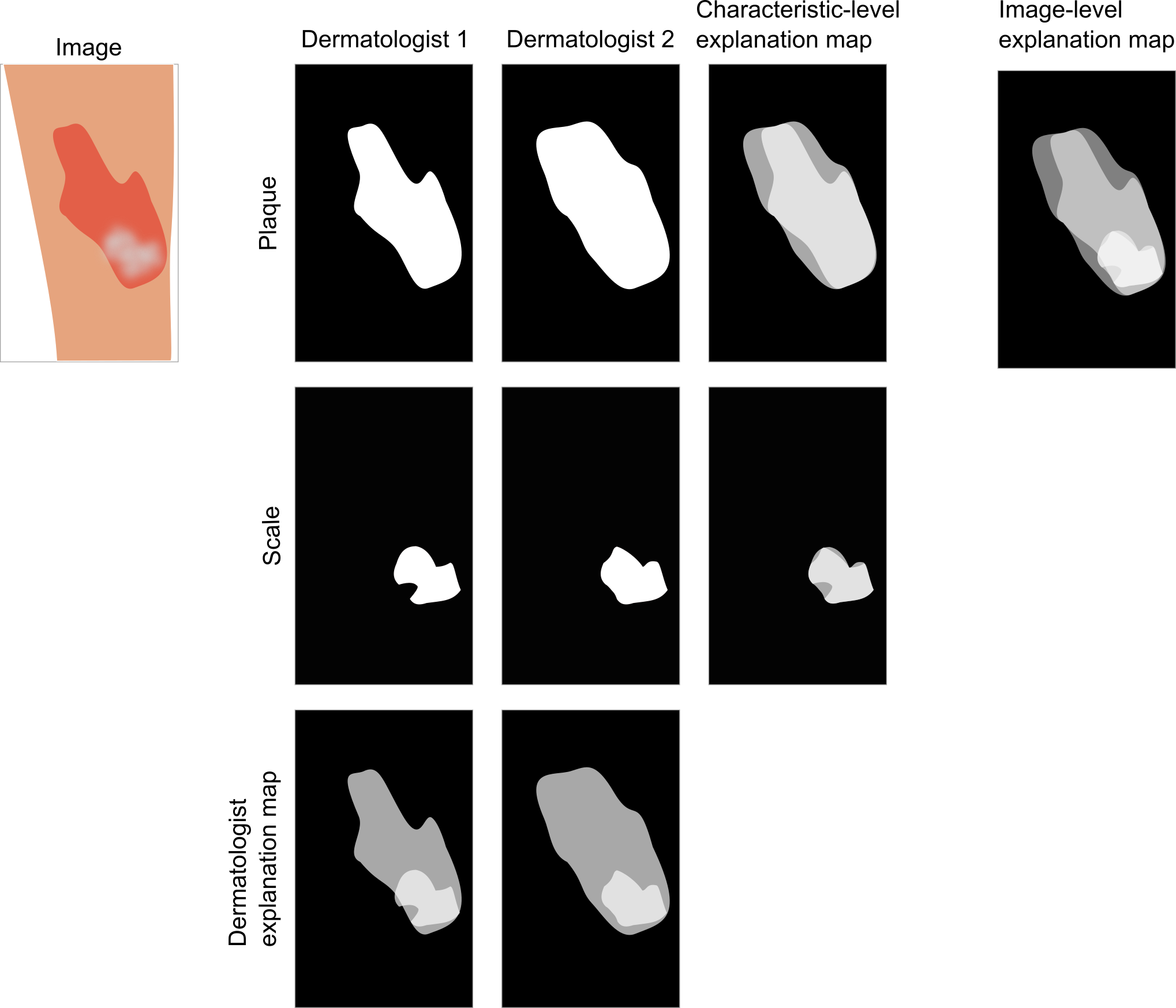}
    \caption{Explanation maps creation example for a psoriasis case evaluated by two dermatologists. Both dermatologists identified plaque and scale as the two characteristics associated with the psoriasis diagnosis, and localized them. By combining the localization maps for each characteristic, we obtain the characteristic-level explanation maps. By combining the localization maps created by each dermatologist, we obtain the individual dermatologist explanation maps. By combining all localization maps, we obtain the image-level explanation map.}
    \label{fig:mask_creation}
\end{figure}

For establishing expert-level explainability performance, we compare the attention masks of each dermatologist with the aggregated fuzzy union of attention masks created by the other seven dermatologists (explanation maps).
More specifically, we define the \emph{image-level explanation maps} as the union of all characteristics segmented by all dermatologists for an image, and the \emph{characteristic-level explanation maps} as the union of all segmentations for each characteristic for an image.
Figure~\ref{fig:mask_creation} illustrates the mask creation process for a psoriasis case.
The ConvNet Grad-CAM attention maps are compared with explanations maps derived from all eight dermatologist evaluations.

These two types of explanation maps offer a way to check whether the ConvNets take into account the entire area selected by dermatologists as important to their decision, and whether they focus on specific characteristics when making their decisions.
To quantify the similarity between the Grad-CAMs and the explanation maps, we compute the F1 score, sensitivity and specificity following their fuzzy implementation defined in~\citep{crum2006}, described as: 


\begin{equation}
    \text{F1 score} = \frac{2\sum_{p\in pixels}{\min(\mathcal{G}_p,\mathcal{E}_p)}}{\sum_{p\in pixels}(\mathcal{G}_p)+\sum_{p\in pixels}(\mathcal{E}_p)},
\end{equation}

\begin{equation}
    \text{Sensitivity} = \frac{\sum_{p\in pixels}{\min(\mathcal{G}_p,\mathcal{E}_p)}}{\sum_{p\in pixels}(\mathcal{S}_p)},
\end{equation}

\begin{equation}
    \text{Specificity} = \frac{\sum_{p\in pixels}{\min(1-\mathcal{G}_p,1-\mathcal{E}_p)}}{\sum_{p\in pixels}(1-\mathcal{E}_p)},
\end{equation}

where $\mathcal{G}$ is the ConvNet-generated Grad-CAM, and $\mathcal{E}$ is the explanation map for a given image.

For characteristics, we report the Grad-CAM sensitivity with regard to the characteristic-level explanation maps.
Specificity and F1 score were considered too stringent, as multiple characteristics can be present and essential for a diagnosis, and an explainable ConvNet must detect all of them to plausibly explain the diagnosis.

\subsubsection{Experimental setup}

From the 22 articles that fulfilled all inclusion criteria, we selected the set of ConvNets to benchmark based on their reproducibility:  we required that all benchmarked ConvNets had been pre-trained on ImageNet due to the limited amount of training data available.
Thus, we exclude architectures that do not have publicly available pre-trained ImageNet weights compatible with the deep learning Keras framework~\citep{chollet2015}, i.e. GoogLeNet~\citep{szegedy2015going}, InceptionV4~\citep{babenko2015aggregating}, MobileNetV3~\citep{howard2019searching}, SENet~\citep{hu2018squeeze}, SE-ResNet~\citep{hu2018squeeze}, SEResNeXT~\citep{hu2018squeeze}, and ShuffleNet~\citep{zhang2018shufflenet}.
Furthermore, as several articles compare different versions of the same architecture (e.g. EfficientNet-B0 through EfficientNet-B7, see Table~\ref{table:final-papers}), we select the smallest version of each architecture for our benchmark to avoid overfitting to the DermXDB dataset.

In the rest of this work, we will focus on the following ConvNets:  DenseNet121~\citep{huang2017densely}, EfficientNet-B0~\citep{tan2019efficientnet}, InceptionResNetV2~\citep{szegedy2017inception}, InceptionV3~\citep{szegedy2016rethinking}, MobileNet~\citep{howard2017mobilenets}, MobileNetV2~\citep{sandler2018mobilenetv2}, NASNetMobile~\citep{zoph2018learning}, ResNet50~\citep{he2016deep}, ResNet50V2~\citep{he2016identity}, VGG16~\citep{simonyan2015very}, and Xception~\citep{chollet2017xception}.

We used the pre-trained weights offered by Keras to initialize the networks in our experiments.
Next, all ConvNets were pre-trained on a proprietary clinical photography skin disease dataset collected by a dermatologist between 2004-2018.
All images included in the dataset were anonymized, and the patients consented to their data being used for research purposes.
More information about the dataset is available in Appendix~Table~\ref{table:pretraining_dataset}.
We performed a hyper-parameter search for each ConvNet, with the values used for experimentation and the validation performance being reported in Appendix~Table~\ref{table:optimised_params} and Appendix~Table~\ref{table:pretraining_diagnosis_performance_f1}, respectively.
We further fine-tuned all ConvNets for 50 epochs with 261 randomly chosen images from the DermXDB dataset.
The remaining 263 images were used as the test set.
Each ConvNet was trained and tested five times.
All results presented in this paper are aggregated over the five test runs. 
All code used for running the experiments is available at~\url{https://github.com/ralucaj/dermx-benchmark}.

\section{Results}
\subsection{Literature review}
\begin{figure}[!t]
    \centering
    \includegraphics[width=0.4\textwidth]{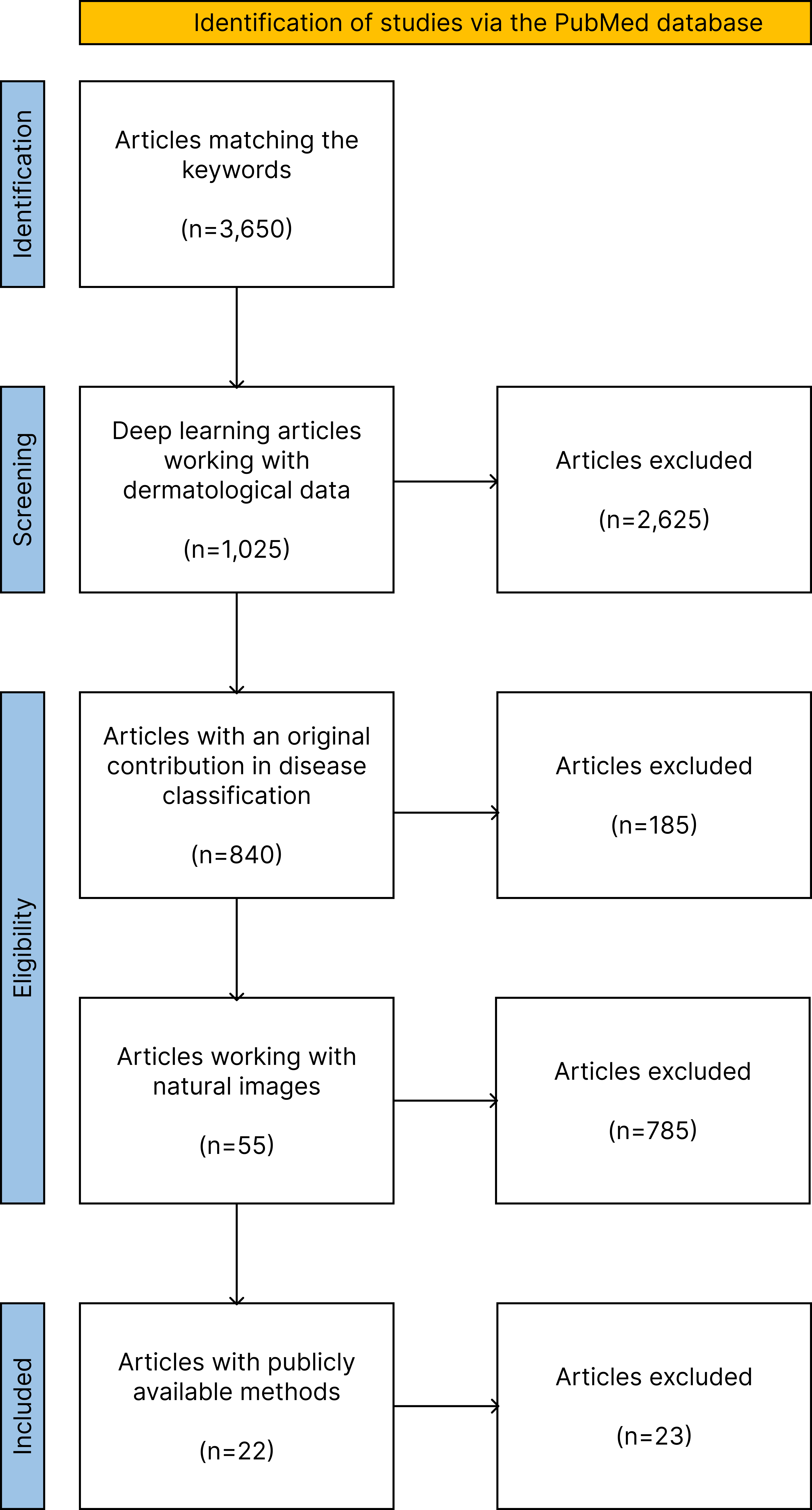}
    \caption{The Preferred Reporting Items for Systematic Reviews and Meta Analyses~(PRISMA) statement flowchart of the performed review process for identifying the benchmarked ConvNet architectures. First, we screened articles to ensure that they were using dermatological data and deep learning methods. Afterwards, we excluded review articles and contributions focused on tasks other than classification, and articles that that used non-photographic image data, e.g. dermoscopy, whole slides. Finally, we excluded articles that used proprietary ConvNets, leading to 22 articles serving as the benchmark basis.}
    \label{fig:exclusion-criteria}
\end{figure}

\begin{figure}[!t]
    \centering
    \includegraphics[width=0.65\textwidth]{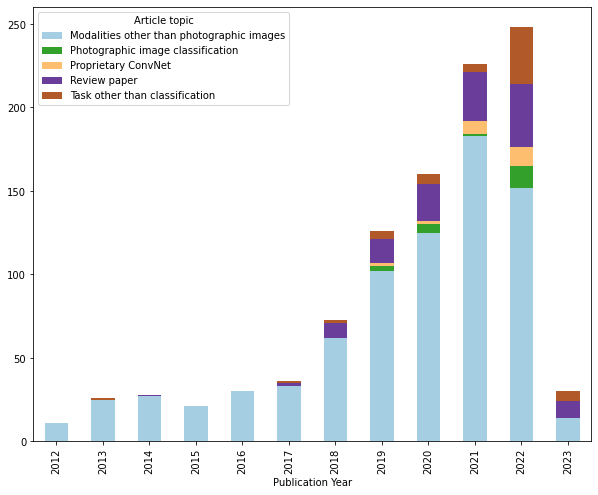}
    \caption{Distribution of retrieved article topics per publication year, based on the search query defined in Table~\ref{table:literature_review_query} (ran on the 20th of February 2023). 2017 marks an explosion in the number of deep learning applications in dermatology, a fact highlighted by the large increase in articles in the subsequent years, and an increase in review articles. Starting 2019, the industrial involvement in this field became apparent due to the increase in proprietary ConvNets. 2019 also marks the first emergence of dermatological applications using photographic imaging. Finally, although classification is still the most common application, other applications are becoming increasingly more researched.}
    \label{fig:papers}
\end{figure}

Figure~\ref{fig:exclusion-criteria} displays the Preferred Reporting Items for Systematic Review and Meta-Analyses statement flowchart of the performed review, while Figure~\ref{fig:papers} illustrates the evolution of articles topics over the years.
Out of the original 3,650 articles, only 22 fulfilled all the inclusion criteria.
Table~\ref{table:final-papers} summarizes the ConvNet architectures, their implementation, and reported performance employed in the final 22 articles selected for benchmarking.

\begin{landscape}
\setlength{\LTcapwidth}{\textwidth}

\begin{longtable}{p{30mm}p{55mm}p{40mm}p{30mm}p{60mm}}


    \caption{Overview of the 22 articles fulfilling all inclusion criteria. All articles use ConvNets for a dermatological classification task using photographic images. Tasks vary between binary or multi-disease diagnosis, disease risk assessment, lesion type classification, and severity assessment.
    \label{table:final-papers}}\\
    
    \toprule
     Publication & ConvNets employed & Task & Data & Performance \\
     \midrule
     \cite{aggarwal2019data} & InceptionV3 & Disease diagnosis on five classes & Open source images and images scraped from Google & 0.66 F1 score, 0.65 sensitivity, 0.91 specificity, 0.67 precision, 0.91 NPV, 0.57 MCC \\ 
     \cite{burlina2019automated} & ResNet50 & Disease diagnosis on four classes & Internet-scraped images & 82.79\% accuracy, 0.76 kappa score \\
     \cite{zhao2019application} & Xception & Skin cancer risk assessment with three classes & Clinical images & 72\% accuracy, 0.92-0.96 ROC AUC, 0.85-0.93 sensitivity, 0.85-0.91 specificity \\
     \cite{burlina2020ai} & ResNet50, ResNet152, InceptionV3, InceptionResNetV2, DenseNet & Disease diagnosis on eight classes & Clinical and other photographic images scraped using Google and Bing & 71.58\% accuracy, 0.70 sensitivity, 0.96 specificity, 0.72 precision, 0.96 NPV, 0.67 kappa, 0.72 F1 score, 0.80 average precision, 0.94 AUC \\
     \cite{chin2020patient} & DenseNet121, VGG16, ResNet50 & Binary skin cancer risk assessment & Smartphone images & 0.83-0.86 AUC, 0.72-0.77 sensitivity, 0.85-0.86 specificity \\
     \cite{han2020augmented} & SENet, SE-ResNet50, VGG19 & Disease classification on 134 classes & Clinical images & 44.8-56.7\% accuracy, 0.94-0.98 AUC\\
     \cite{liu2020deep} & InceptionV4 & Disease diagnosis on 26 classes & Clinical images & 66\% accuracy, 0.56 sensitivity\\
     \cite{zhao2020smart} & DenseNet121, Xception, InceptionV3, InceptionResNetV2 & Binary psoriasis classification & Clinical images & 96\% accuracy, 0.95-0.98 AUC, 0.96-0.97 specificity, 0.83-0.95 sensitivity\\
     \cite{wu2021deep} & SEResNeXt, SE-ResNet, InceptionV3 & Disease diagnosis on five classes & Clinical images & 0.96-0.97 AUC, 90-91\% accuracy, 0.90-0.93 sensitivity, 0.90 specificity\\
     \cite{aggarwal2022artificial} & InceptionResNetV2 & Disease diagnosis on four classes & Clinical images & 0.60-0.82 sensitivity, 0.60-0.82 specificity, 0.33-0.93 precision, 0.33-0.93 NPV, 0.43-0.84 F1 score \\
     \cite{ba2022convolutional} & EfficientNet-B3 & Disease diagnosis on 10 classes & Clinical images & 78.45\% accuracy, 0.73 kappa \\
     \cite{hossain2022exploring} & VGG16, VGG19, ResNet50, ResNet101, ResNet50V2, ResNet101V2, InceptionV3, InceptionV4, InceptionResNetV2, Xception, DenseNet121, DenseNet169, DenseNet201, MobileNetV2, \mbox{MobileNetV3Small}, \mbox{MobileNetV3Large}, NASNetMobile, \mbox{EfficientNet-B0} through EfficientNet-B5 & Binary Lyme disease classification & Smartphone images & 61.42-84.42\% accuracy, 0.72-0.90 sensitivity, 0.50-0.81 specificity, 0.61-0.83 precision, 0.63-0.87 NPV, 0.23-0.69 MCC, 0.22-0.69 Cohen's kappa, 1.46-4.70 positive likelihood ratio, 0.14-0.55 negative likelihood ratio, 0.66-0.0.85 F1 score, 0.65-0.92 AUC\\
     \cite{husers2022automatic} & MobileNet & Binary wound maceration classification & Clinical images & 69\% accuracy, 0.69 sensitivity, 0.67 precision\\
     \cite{liu2022pressure} & InceptionResNetV2 & Ulcer characteristic diagnosis on two and three classes & Clinical images & 71.2-99.4\% accuracy, 0.68-0.99 sensitivity, 0.71-1.00 precision, 0.70-0.94 F1 score \\
     \cite{malihi2022automatic} & Xception & Binary wound type classification & Clinical images & 67-83\% accuracy, 0.65-0.94 sensitivity, 0.70-0.75 specificity, 0.65-0.75 precision, 0.70-0.85 F1 score \\ 
     \cite{munthuli2022extravasation} & DenseNet121 & Skin lesion severity classification with five classes & Smartphone images & 0.43-0.91 sensitivity, 0.80-0.98 specificity, 0.50-0.87 F1 score \\
     \cite{ni2022deep} & DenseNet121, ResNet50 & Radiation dermatitis severity classification on four classes & Clinical images & 83\% accuracy, 0.74-1.00 F1 score \\
     \cite{roy2022does} & ResNet101 & Disease diagnosis on 26 classes & Clinical images & 62.6\% - 75.6\% accuracy, 69.3-81.8 AUPR\\
     \cite{sahin2022human} & ResNet18, GoogleNet, EfficientNet-B0, NASNetMobile, ShuffleNet, \mbox{MobileNetV2} & Binary monkeypox classification & Smartphone images & 73.33-91.11\% accuracy \\
     \cite{xia2022lesion} & ResNet50 & Binary skin cancer classification & Smartphone images & 0.77-0.82 AUC, 0.76-0.79 AP \\
     \cite{zhou2022background} & ResNet50 & Disease diagnosis on three classes & Clinical images & 0.32 error rate, 0.68 sensitivity, 0.69 precision, 0.68 F1 score \\
     \cite{zhang2022acne} & ResNet50 & Acne severity classification with three classes & Clinical images & 74\% accuracy \\
     \bottomrule     
\end{longtable}
\end{landscape}

\subsection{Diagnosis results}
\begin{table}[t!]
    \setlength{\tabcolsep}{7pt} 
    \renewcommand{\arraystretch}{1.5} 
    \centering
    \caption{Diagnostic performance of the ConvNets~(average $\pm$ standard deviation across five runs) and dermatologists~(average $\pm$ standard deviation across eight experts) using F1-score, split by diagnosis. Several ConvNets achieve expert-level per-disease diagnosis performance, in actinic keratosis, seborrheic dermatitis, and viral warts~(in \textbf{bold}), although none reach the same performance for acne, psoriasis, and vitiligo.}
    \begin{tabular}{lrrrrrr}
    \toprule
         & 
         \multicolumn{1}{p{10pt}}{\centering\textbf{Acne}} & 
         \multicolumn{1}{p{10pt}}{\centering \textbf{Actinic} \\ \textbf{keratosis}} & 
         \multicolumn{1}{p{10pt}}{\centering\textbf{Psoriasis}} & 
         \multicolumn{1}{p{10pt}}{\centering \textbf{Seborrheic} \\ \textbf{dermatitis}} & 
         \multicolumn{1}{p{10pt}}{\centering \textbf{Viral} \\ \textbf{warts}}  & 
         \multicolumn{1}{p{10pt}}{\centering \textbf{Vitiligo}} \\
    \midrule
        \textbf{ConvNets} &  &  &  &  &  &  \\
        DenseNet121 & $0.80 \pm 0.02$ & $\mathbf{0.63 \pm 0.08}$ & $0.66 \pm 0.01$ & $\mathbf{0.69 \pm 0.03}$ & $\mathbf{0.88 \pm 0.03}$ & $0.74 \pm 0.03$  \\
        EfficientNet-B0 & $0.72 \pm 0.03$ & $0.53 \pm 0.10$ & $0.60 \pm 0.06$ & $\mathbf{0.57 \pm 0.08}$ & $0.80 \pm 0.07$ & $0.66 \pm 0.02$  \\
        InceptionV3 & $0.77 \pm 0.02$ & $\mathbf{0.57 \pm 0.11}$ & $0.60 \pm 0.02$ & $0.54 \pm 0.03$ & $0.77 \pm 0.04$ & $0.73 \pm 0.05$  \\
        InceptionResNetV2 & $0.73 \pm 0.02$ & $0.52 \pm 0.10$ & $0.53 \pm 0.05$ & $0.56 \pm 0.05$ & $0.69 \pm 0.03$ & $0.53 \pm 0.12$  \\
        MobileNet & $0.72 \pm 0.06$ & $\mathbf{0.55 \pm 0.19}$ & $0.51 \pm 0.14$ & $\mathbf{0.57 \pm 0.06}$ & $0.68 \pm 0.06$ & $0.56 \pm 0.10$  \\
        MobileNetV2 & $0.56 \pm 0.07$ & $0.23 \pm 0.09$ & $0.31 \pm 0.08$ & $0.46 \pm 0.05$ & $0.63 \pm 0.07$ & $0.48 \pm 0.14$  \\
        NASNetMobile & $0.50 \pm 0.05$ & $0.33 \pm 0.12$ & $0.42 \pm 0.07$ & $0.43 \pm 0.05$ & $0.55 \pm 0.11$ & $0.51 \pm 0.05$  \\
        ResNet50 & $0.77 \pm 0.04$ & $\mathbf{0.53 \pm 0.17}$ & $0.61 \pm 0.03$ & $\mathbf{0.61 \pm 0.19}$ & $0.79 \pm 0.02$ & $0.61 \pm 0.07$  \\
        ResNet50V2 & $0.76 \pm 0.04$ & $\mathbf{0.62 \pm 0.07}$ & $0.59 \pm 0.01$ & $0.57 \pm 0.01$ & $0.76 \pm 0.01$ & $0.75 \pm 0.05$  \\
        VGG16 & $0.70 \pm 0.05$ & $\mathbf{0.62 \pm 0.03}$ & $0.59 \pm 0.03$ & $\mathbf{0.49 \pm 0.15}$ & $0.71 \pm 0.03$ & $0.62 \pm 0.07$  \\
        Xception & $0.80 \pm 0.04$ & $\mathbf{0.64 \pm 0.07}$ & $0.70 \pm 0.02$ & $\mathbf{0.60 \pm 0.03}$ & $0.81 \pm 0.04$ & $0.81 \pm 0.05$  \\ [5pt]
        \textbf{Dermatologists} &  &  &  &  &  &  \\
        Average & $0.95 \pm 0.02$ & $0.79 \pm 0.14$ & $0.85 \pm 0.06$ & $0.72 \pm 0.09$ & $0.93 \pm 0.05$ & $0.96 \pm 0.03$ \\
    \bottomrule
    \end{tabular}
    \label{table:diagnosis_performance_f1}
\end{table}

Table~\ref{table:diagnosis_performance_f1} provides an overview of the diagnostic performance of the networks and that of the dermatologists on average in terms of F1~score. 
As can be seen from the table, although several ConvNets achieve expert-level performance when diagnosing actinic keratosis, seborrheic dermatitis, and viral warts, none of them achieve overall expert-level performance. 
ConvNets follow the trend also seen in dermatologists of having difficulties correctly diagnosing actinic keratosis and seborrheic dermatitis, while the diagnosis of acne and viral warts displays higher performance.
Similar trends can be observed for the sensitivity and specificity performance, as seen in Appendix~Table~\ref{table:diagnosis_performance_sensitivity} and Appendix~Table~\ref{table:diagnosis_performance_specificity}, respectively.

\subsection{Explainability results}

\begin{figure}[tb]
    \centering
    \includegraphics[width=0.65\textwidth]{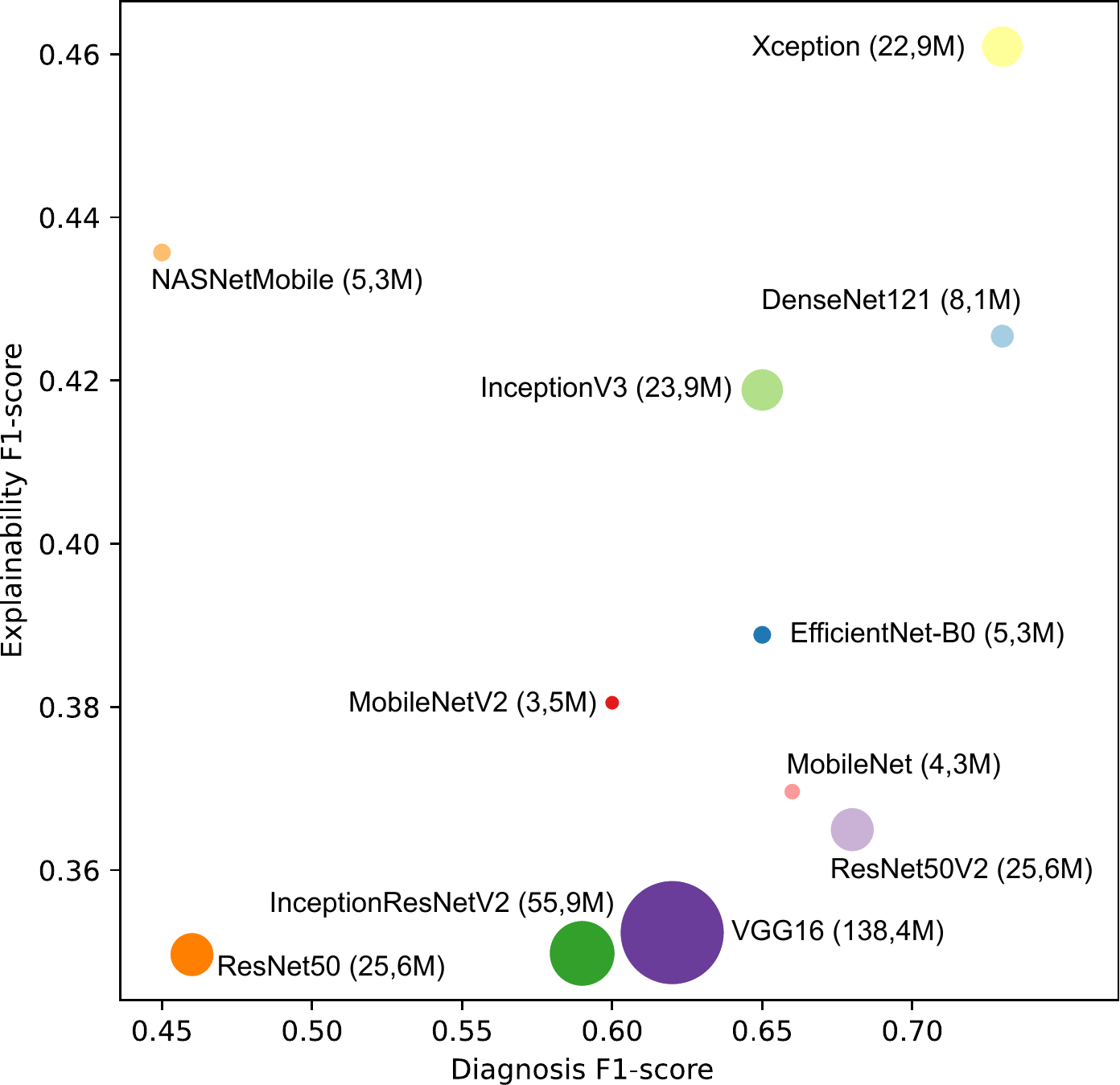}
    \caption{ConvNet explainability as a function of ConvNet performance and their number of parameters. Xception displays both the highest performance and image-level explainability, while ResNet50 performs poorly in both criteria.}
    \label{fig:explainability_vs_performance}
\end{figure}

\begin{table}[t!]
    \setlength{\tabcolsep}{10pt} 
    \renewcommand{\arraystretch}{1.5} 
    \centering
    \caption{Explainability performance in terms of the image-level Grad-CAM evaluation for the ConvNets~(average $\pm$ standard deviation across five runs), and an explanation map evaluation dermatologists~(average $\pm$ standard deviation across eight experts). Older ConvNets, such as ResNet50, ResNet50V2, and VGG16, have lower performance than most other modern ConvNets. Two networks achieve expert-level sensitivity scores, and one achieves expert-level specificity~(in \textbf{bold}).}

    \begin{tabular}{lrrrr}
    \toprule
         & \textbf{F1 score} & \textbf{Sensitivity} & \textbf{Specificity} \\
    \midrule
        \textbf{ConvNets}\\
        DenseNet121 & $0.43 \pm 0.01$ & $\mathbf{0.61 \pm 0.01}$ & $0.78 \pm 0.00$ \\
        EfficientNet-B0 & $0.39 \pm 0.01$ & $0.52 \pm 0.00$ & $0.82 \pm 0.00$ \\
        InceptionV3 & $0.42 \pm 0.01$ & $0.56 \pm 0.01$ & $0.82 \pm 0.01$ \\
        InceptionResNetV2 & $0.35 \pm 0.01$ & $0.40 \pm 0.01$ & $0.87 \pm 0.01$ \\
        MobileNet & $0.37 \pm 0.02$ & $0.50 \pm 0.01$ & $0.85 \pm 0.01$ \\
        MobileNetV2 & $0.38 \pm 0.02$ & $0.49 \pm 0.02$ & $0.87 \pm 0.01$ \\
        NASNetMobile & $0.44 \pm 0.00$ & $\mathbf{0.62 \pm 0.00}$ & $0.81 \pm 0.00$ \\
        ResNet50 & $0.35 \pm 0.01$ & $0.42 \pm 0.03$ & $0.84 \pm 0.01$ \\
        ResNet50V2 & $0.37 \pm 0.01$ & $0.38 \pm 0.01$ & $\mathbf{0.91 \pm 0.00}$ \\
        VGG16 & $0.35 \pm 0.01$ & $0.40 \pm 0.01$ & $0.86 \pm 0.01$ \\
        Xception & $0.46 \pm 0.01$ & $0.56 \pm 0.00$ & $0.88 \pm 0.01$ \\ [5pt]
        \textbf{Dermatologists}\\
        Average & $0.66 \pm 0.03$ & $0.67 \pm 0.07$ & $0.93 \pm 0.03$\\
    \bottomrule
    \end{tabular}
    \label{table:per_image_explainability}
\end{table}

Table~\ref{table:per_image_explainability} shows the image-level explainability results for each of the benchmarked ConvNets, while Figure~\ref{fig:explainability_vs_performance} shows the relationship between ConvNet diagnosis performance, image-level explainability, and number of parameters.
Xception scores the highest on the image-level Grad-CAM F1 score, while InceptionResNetV2, ResNet50, and VGG16 have the lowest performance.
DenseNet121 and NASNetMobile report expert-level sensitivity scores, while ResNet50V2 achieves expert-level performance in specificity.

\begin{figure}[t!]
    \centering
    \includegraphics[width=0.85\textwidth]{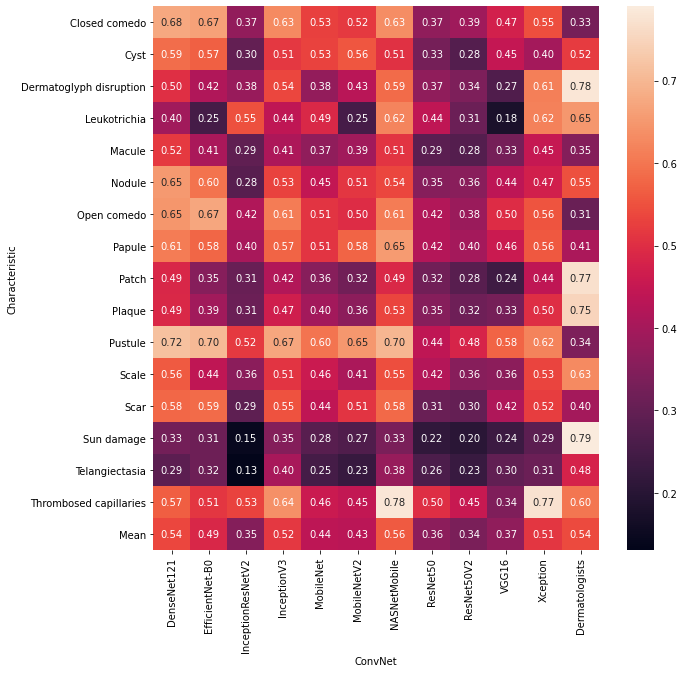}
    \caption{Explainability performance in terms of characteristic-level Grad-CAM sensitivity for the ConvNets~(averaged across five runs) and dermatologists~(averaged across eight experts). NASNetMobile and Xception outperform expert level in seven characteristics, while no ConvNet achieves expert-level performance in eight characteristics.}
    \label{fig:characteristic_performance_comparative}
\end{figure}

Looking at the characteristic-level sensitivity depicted in Figure~\ref{fig:characteristic_performance_comparative}, NASNetMobile and DenseNet121 achieve the highest overall performance. 
InceptionResNetV2, ResNet50, ResNet50V2, and VGG16 report the lowest scores.
All ConvNets outperform dermatologists in closed comedo, open comedo, and pustule.
The opposite is true for dermatoglyph disruption, leukotrichia, patch, plaque, scale, sun damage, and telangiectasia -- no ConvNet reaches expert-level.

\begin{figure}[t!]
    \centering
    \includegraphics[width=0.77\textwidth]{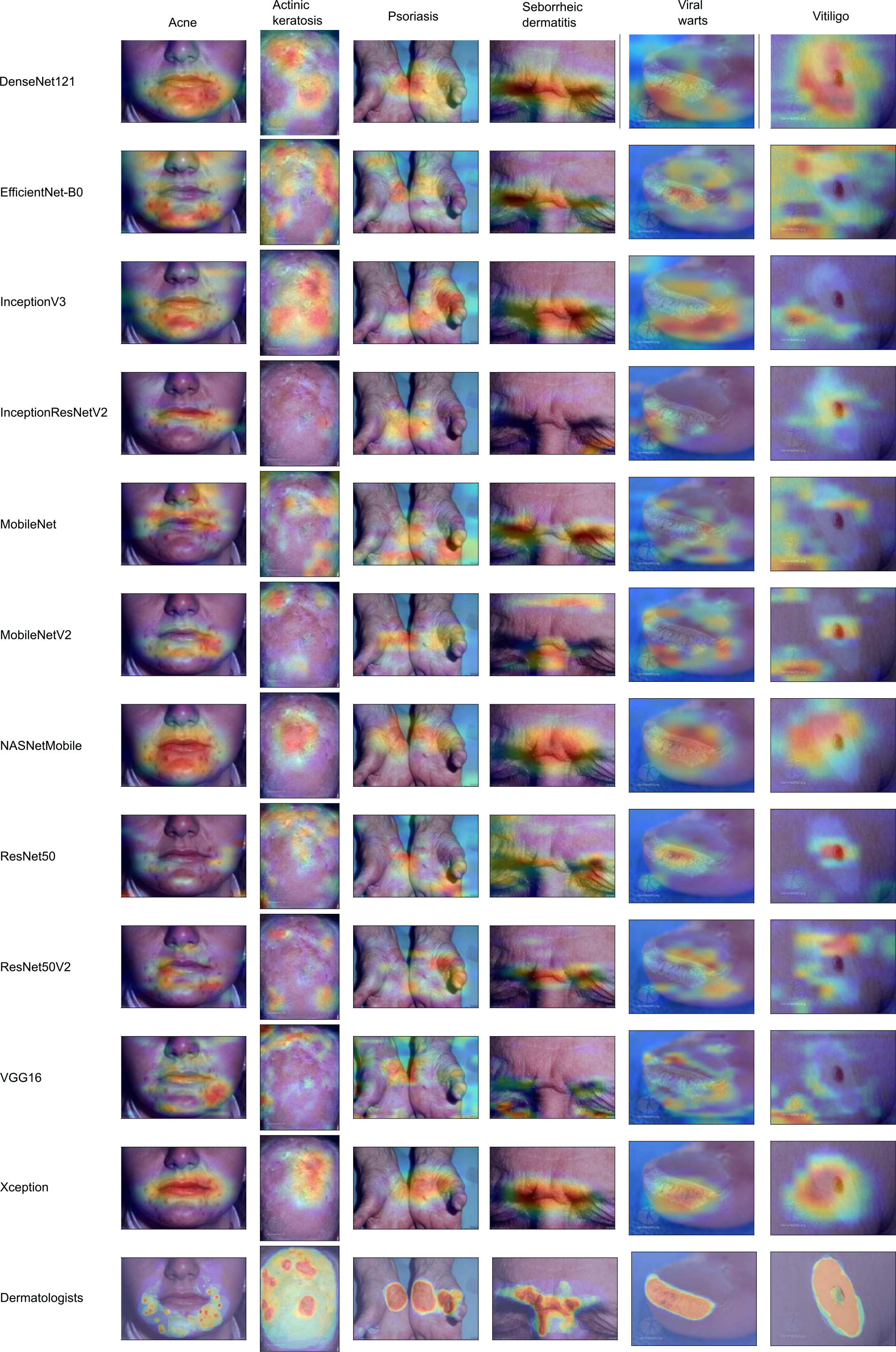}
    \caption{Example of Grad-CAM outputs for six images that were correctly diagnosed by all ConvNets. Older ConvNets, such as VGG16, ResNet50, ResNet50V2, and InceptionResNetV2, tend to focus on a single, highly indicative lesion rather than the whole affected region. More modern ConvNets, such as NASNetMobile, Xception, and EfficientNet, focus on the entire affected area. Some ConvNets overfitted during training, and focus on the watermark when diagnosing vitiligo.}
    \label{fig:gradcam_examples}
\end{figure}

Figure~\ref{fig:gradcam_examples} illustrates the differences in Grad-CAMs between the benchmarked ConvNets. 
Older ConvNet architectures, such as VGG16, InceptionResNetV2, ResNet50, and ResNet50V2, tend to focus on small areas that contain characteristics relevant for the diagnosis, e.g. focusing on a single plaque in the psoriasis diagnosis example, while more modern ConvNets pay attention to the entire area covered by diagnosis-relevant lesions.
Several ConvNets, namely EfficientNet-B0, MobileNet, MobileNetV2, and VGG16 seem to have overfit on the training set, focusing on the watermark rather than the image itself when diagnosing the vitiligo case.

\section{Discussion}
\subsection{Literature review}
ConvNets have become a default approach when it comes to automated diagnosis using images, aligned with the rise of the deep learning methodology for vision recognition. 
The continuous breakthroughs in diagnostic performance across a wide variety of medical imaging modalities and disorders have made automated diagnosis as close to integration with practice as ever. 
In dermatology, the diagnosis performance has achieved that of the expert raters as early as 2017 with a seminal work of~\cite{esteva2017dermatologist} that  disrupted the research field and set the trend that still persists, as can be seen through the trends of the continuous growth outlined in~Figure~\ref{fig:papers}.
The increased interest of industrial entities that started in 2019, illustrated in Figure~\ref{fig:papers} by the increase in proprietary methods, is further highlighted by the large number of dermatology-oriented med-tech companies relying on machine learning for their products.
Year 2019 also marks the year when research groups began investigating photographic images as a primary modality for diagnosing skin conditions, meaning the rise of machine learning solutions to assist teledermatology. 

The potential of using ConvNets to streamline dermatological tasks is underlined by the diversity of tasks being solved in the retrieved articles.
Classification was the first methodology to be approached, with applications in disease diagnosis, risk assessment, lesion type classification, lesion characteristics identification, and disease severity assessment.
Segmentation and natural language processing applications are also gaining more traction, as shown by the constant increase in non-classification tasks in Figure~\ref{fig:papers}.

However, this potential has not yet translated into the much-needed transformation of the clinical practice. 
In part, this is due to regulatory challenges which are often faced due to the limited generalizability and lack of explainability of the methods~\citep{kelly2019key}.
By benchmarking the diagnosis and explainability performance of ConvNets, we both enable a comparison among the methods, as well as help the identification gaps between the current state-of-the-art and the clinical practice.

\subsection{Diagnosis benchmark}
The direct comparison of the diagnostic performance is not possible using reported values from the literature not only due to variability in the choice of the metrics, but more importantly due to the variance in the number of classes and the differences in the datasets used for training and validation~(Table~\ref{table:final-papers}).
By reformulating the task to the diagnosis of six disease classes, utilizing the same initialization, pre-training, and hyperparameter optimization search strategy, and training and validating on the common database, this benchmark minimizes the performance variability related to such implementation details. 

We found considerable variability among the diagnostic performance values, with the average F1 scores ranging from 0.50 to 0.80 for acne, from 0.23 to 0.64 for actinic keratosis, from 0.31 to 0.70 for psoriasis, from 0.43 to 0.69 for seborrheic dermatitis, from 0.55 to 0.88 for viral warts, and from 0.51 to 0.81 for vitiligo.
These values were aligned with the diagnostic complexity of the diseases as expressed by the performance of the dermatologists, averaging 0.95 for acne, 0.79 for actinic keratosis, 0.85 for psoriasis, 0.72 for seborrheic dermatitis, 0.93 for viral warts, and 0.96 for vitiligo.
As such, none of the ConvNets achieved the average dermatologist performance, although there were multiple instances of ConvNets reaching the range of the expert performance for a specific disease~(see~Table~\ref{table:diagnosis_performance_f1}).
The majority of the benchmarked ConvNets achieved expert level for diagnosis of actinic keratosis and seborrheic dermatitis: seven and six out of 11, respectively.
This further confirms the similarity of ConvNet performance with respect to the dermatologists: most ConvNets display a similar difficulty in diagnosing actinic keratosis and seborrheic dermatitis as the eight dermatologists, and a similar ease of diagnosing acne and viral warts.

\subsection{Explainability benchmark}
While diagnostic performance is recognized as critical for the generalizability of ConvNets, the explainability performance validation has been generally approached as an optional, qualitative, post-hoc analysis.
One of the key challenges faced by researchers trying to implement a more objective validation of explainability is linking the human-approachable explanations with those feasible for ConvNets.
With the use of the labels for dermatological diagnosis explainability available from the recently released DermXDB~dataset, our benchmark is quantitative as well as predefined.
Thus, we avoid potential biases and limitations stemming from machine learning experts with little domain knowledge performing a visual, qualitative evaluation of Grad-CAMs~\citep{tschandl2020human}.

The image-level explainability analysis shows that no ConvNet reaches the same F1 score as the dermatologists, although several ConvNets achieve expert-level sensitivity or specificity.
Different ConvNets show different patterns of explanation behaviour (Figure~\ref{fig:gradcam_examples}): some tend to focus on smaller areas that are highly indicative of the target diagnosis, while others tend to focus on the entire affected area. 
Extensive user tests with both experts and patients would enable us to learn which of the two options is preferred as an explanation: a single, classical lesion descriptive of the diagnosis, or highlighting the entire affected area.

From a characteristic-level sensitivity perspective, most ConvNets outperform the average dermatologist performance in characteristics smaller than 1cm in diameter~\citep{nast20162016}. 
For larger characteristics, although NASNetMobile and Xception approach expert-level, no ConvNet exceeds it.
The relationship between diseases and their characteristics is visible in the characteristic-level ConvNet explainability: most ConvNets report high sensitivity on characteristics often associated with acne and viral warts (e.g. closed and open comedones, papules, and thrombosed capillaries), while reporting a lower performance on characteristics associated with actinic keratosis and seborrheic dermatitis (e.g. plaque, sun damage, and patch).  
Characteristic-level explainability may be more relevant for use cases where identifying the differentiating factor between different diseases is the most important component for garnering trust.  

These result suggests that while ConvNets have the potential to produce human-approachable explanations, more work is necessary to fully achieve expert-level performance.
Part of the necessary work is the creation of additional user-derived explainability datasets that enable quantitative analyses on a ConvNet's explainability within a domain.
A component of this is performing extensive user tests to identify the explainability expectations of an application's end users.  
From a machine learning perspective, more research must be devoted to the creation of instrinsically explainable ConvNets, rather than relying solely on post-hoc explanation methods. 
Such a ConvNet must be aligned with the explainability requirements of its task and its users: a psoriasis diagnosis ConvNet aimed at dermatologists might require high characteristic-level explainability to offer a constrative explanation against a possible differential diagnosis of atopic dermatitis, while the same ConvNet aimed at patients might require high image-level explainability to reassure the patient that all aspects of their condition are taken into consideration. 


\subsection{Limitations and future work}
Our work has a few limitations.
First, the original DermXDB dataset contains little information about the gender, age, and ethnicity of the subjects, leading to difficulties in performing an in-depth bias analysis of our benchmark.
Second, the small size of the dataset limits the training capabilities of our benchmark, which may underestimate the performance of the larger ConvNets.

In future work, we plan on expanding this benchmark by using more explainability methods, such as saliency maps and LIME, to also create a benchmark of explainability methods and their performance compared to that of dermatologists.
Additionally, with the increased popularity of visual transformers~\citep{khan2022transformers}, an analysis of their Grad-CAM explainability would be of interest to the research world.


\section{Conclusions}
In this paper, we performed a systematic literature review to identify the most used ConvNet architectures for the diagnosis of skin diseases from photographic images.
We benchmarked the 11 identified ConvNets on DermXDB, a skin disease explainability dataset.
Xception stands out as a highly explainable ConvNet, although NASNetMobile outperforms it on characteristic-level sensitivity. 
Our findings highlight the importance of explainability benchmarking, and will hopefully motivate additional studies within the field of quantitative evaluations for explainability.

\section*{Acknowledgments} 
Funding: RJ's work was supported in part by the Danish Innovation Fund under Grant 0153-00154A. OW's work was funded in part by the Novo Nordisk Foundation through the Center for Basic Machine Learning Research in Life Science (NNF20OC0062606). OW acknowledges support from the Pioneer Centre for AI, DNRF grant number P1.

\typeout{}

\bibliographystyle{unsrtnat}
\bibliography{main}

\clearpage
\onecolumn

\appendix
\section*{Appendix}
\renewcommand{\thesection}{A\arabic{section}}
\renewcommand{\thefigure}{A\arabic{figure}}
\renewcommand\thetable{A\arabic{table}}
\setcounter{section}{0} 
\setcounter{figure}{0} 
\setcounter{table}{0}  
Table~\ref{table:pretraining_dataset} presents statistics for the proprietary clinical dataset used in the hyper-parameter search and the pre-training step. 
Table~\ref{table:optimised-params} shows the best performing list of parameters identified for each ConvNet. 
The search space consisted of the following values for each hyperparameter:
\begin{itemize}
    \item{\textbf{Rotation:} 10, 20}
    \item{\textbf{Shear:} 0.00, 0.25, 0.50}
    \item{\textbf{Zoom:} 0.25, 0.5}
    \item{\textbf{Brightness ranges:} 0.00-0.50, 0.00-0.25, 0.50-1.00, 0.50-1.50, 0.75-1.25}
    \item{\textbf{Learning rate:} 0.01, 0.001, 0.0001}
    \item{\textbf{Last fixed layer:} last convolutional layer, second to last convolutional block}
    \item{\textbf{Epochs:} 10, 25, 50, 75}
\end{itemize}

\begin{table}[h!]
    \setlength{\tabcolsep}{7pt} 
    \renewcommand{\arraystretch}{1.5} 
    \centering
    \caption{Dataset statistics for the proprietary pre-training clinical dataset.}

    \label{table:pretraining_dataset}
    \begin{tabular}{lrr}
    \toprule
        Diagnosis & Training & Validation \\
    \midrule
        Acne & 832 & 245 \\
        Actinic keratosis & 132 & 33 \\
        Psoriasis & 771 & 204 \\
        Seborrheic dermatitis & 88 & 25 \\
        Viral warts & 509 & 97 \\
        Vitiligo & 141 & 37 \\
    \bottomrule
    \end{tabular}
\end{table}

\begin{table}[h!]
    \setlength{\tabcolsep}{7pt} 
    \renewcommand{\arraystretch}{1.5} 
    \centering
    \caption{Optimal list of hyperparameters for each ConvNet, as identified after a hyper-parameter search.}

    \label{table:optimised_params}
    \begin{tabular}{lrrrrrlr}
    \toprule
        ConvNet & Rotation & Shear & Zoom & Brightness & Learning rate & Last fixed layer  & Epochs \\
    \midrule
        DenseNet121 & 20 & 0.50 & 0.50 & [0.50, 1.50] & 0.0001 & conv5\_block14\_concat & 75 \\
        EfficientNet-B0 & 20 & 0.25 & 0.50 & [0.50, 1.50] & 0.0001 & block6d\_add & 50 \\
        InceptionV3 & 20 & 0.50 & 0.50 & [0.50, 1.50] & 0.001 & activation288 & 50 \\
        InceptionResNetv2 & 20 & 0.25 & 0.50 & [0.75, 1.25] & 0.0001 & block8\_9\_ac & 50 \\
        MobileNet & 10 & 0.50 & 0.50 & [0.50, 1.50] & 0.0001 & conv\_pw\_12\_relu & 50 \\
        MobileNetV2 & 10 & 0.25 & 0.50 & [0.50, 1.50] & 0.0001 & block\_15\_add & 75 \\
        NASNetMobile & 20 & 0.25 & 0.50 & [0.50, 1.00] & 0.0001 & normal\_concat\_11 & 75 \\
        ResNet50 & 20 & 0.50 & 0.50 & [0.50, 1.50] & 0.0001 & conv5\_block3\_out & 50 \\
        ResNet50V2 & 20 & 0.25 & 0.25 & [0.50, 1.00] & 0.001 & post\_relu & 75 \\
        VGG16 & 10 & 0.00 & 0.25 & [0.50, 1.00] & 0.01 & block5\_pool & 75 \\
        Xception & 10 & 0.25 & 0.50 & [0.50, 1.50] & 0.001 & block14\_sepconv2\_act & 50\\
    \bottomrule
    \end{tabular}
    \label{table:optimised-params}
\end{table}

\begin{table}[h!]
    \setlength{\tabcolsep}{7pt} 
    \renewcommand{\arraystretch}{1.5} 
    \centering
    \caption{Diagnostic performance of ConvNets in terms macro F1-score, sensitivity, and specificity on the validation subset of the proprietary clinical dataset~(average $\pm$ standard deviation across five runs).}
    \begin{tabular}{lrrrrrr}
    \toprule
        ConvNet & F1-score & Sensitivity & Specificity \\
    \midrule
        DenseNet121 & $0.80 \pm 0.01$ & $0.79 \pm 0.01$ & $0.98 \pm 0.00$ \\
        EfficientNet-B0 & $0.77 \pm 0.01$ & $0.78 \pm 0.01$ & $0.97 \pm 0.00$ \\
        InceptionV3 & $0.76 \pm 0.02$ & $0.74 \pm 0.02$ & $0.96 \pm 0.00$ \\
        InceptionResNetV2 & $0.73 \pm 0.02$ & $0.73 \pm 0.02$ & $0.97 \pm 0.00$ \\
        MobileNet & $0.72 \pm 0.02$ & $0.71 \pm 0.02$ & $0.96 \pm 0.00$ \\
        MobileNetV2 & $0.72 \pm 0.03$ & $0.73 \pm 0.02$ & $0.96 \pm 0.00$  \\
        NASNetMobile & $0.67 \pm 0.04$ & $0.64 \pm 0.02$ & $0.95 \pm 0.01$ \\
        ResNet50 & $0.70 \pm 0.01$ & $0.68 \pm 0.02$ & $0.96 \pm 0.00$ \\
        ResNet50V2 & $0.76 \pm 0.01$ & $0.75 \pm 0.02$ & $0.96 \pm 0.00$ \\
        VGG16 & $0.66 \pm 0.03$ & $0.67 \pm 0.01$ & $0.95 \pm 0.00$ \\
        Xception & $0.82 \pm 0.03$ & $0.82 \pm 0.02$ & $0.97 \pm 0.00$ \\
    \bottomrule
    \end{tabular}
    \label{table:pretraining_diagnosis_performance_f1}
\end{table}


\begin{table}[h!]
    \setlength{\tabcolsep}{7pt} 
    \renewcommand{\arraystretch}{1.5} 
    \centering
    \caption{Diagnostic performance of ConvNets~(average $\pm$ standard deviation across five runs) and dermatologists~(average $\pm$ standard deviation across eight dermatologists) in terms of sensitivity on the DermXDB holdout set, split by diagnosis. Several ConvNets achieve expert-level sensitivity on multiple classes~(in~\textbf{bold}).}
    \begin{tabular}{lrrrrrr}
    \toprule
         & 
         \multicolumn{1}{p{10pt}}{\centering\textbf{Acne}} & 
         \multicolumn{1}{p{10pt}}{\centering \textbf{Actinic} \\ \textbf{keratosis}} & 
         \multicolumn{1}{p{10pt}}{\centering\textbf{Psoriasis}} & 
         \multicolumn{1}{p{10pt}}{\centering \textbf{Seborrheic} \\ \textbf{dermatitis}} & 
         \multicolumn{1}{p{10pt}}{\centering \textbf{Viral} \\ \textbf{warts}}  & 
         \multicolumn{1}{p{10pt}}{\centering \textbf{Vitiligo}} \\
    \midrule
        \textbf{ConvNets}\\
        DenseNet121 & $0.85 \pm 0.03$ & $\mathbf{0.52 \pm 0.10}$ & $0.71 \pm 0.03$ & $\mathbf{0.76 \pm 0.05}$ & $\mathbf{0.91 \pm 0.03}$ & $0.66 \pm 0.03$ \\
        EfficientNet-B0 & $0.71 \pm 0.08$ & $\mathbf{0.43 \pm 0.09}$ & $0.63 \pm 0.07$ & $\mathbf{0.64 \pm 0.13}$ & $0.66 \pm 0.05$ & $\mathbf{0.84 \pm 0.13}$ \\
        InceptionV3 & $0.83 \pm 0.06$ & $\mathbf{0.55 \pm 0.15}$ & $0.62 \pm 0.08$ & $\mathbf{0.54 \pm 0.13}$ & $\mathbf{0.70 \pm 0.10}$ & $0.73 \pm 0.12$\\
        InceptionResNetV2 & $0.70 \pm 0.06$ & $\mathbf{0.41 \pm 0.12}$ & $0.62 \pm 0.06$ & $\mathbf{0.67 \pm 0.11}$ & $0.43 \pm 0.12$ & $0.74 \pm 0.07$ \\
        MobileNet & $0.76 \pm 0.07$ & $\mathbf{0.48 \pm 0.22}$ & $0.60 \pm 0.26$ & $\mathbf{0.66 \pm 0.08}$ & $0.47 \pm 0.17$ & $0.68 \pm 0.07$ \\
        MobileNetV2 & $\mathbf{0.88 \pm 0.08}$ & $0.14 \pm 0.06$ & $0.21 \pm 0.10$ & $\mathbf{0.63 \pm 0.22}$ & $0.33 \pm 0.14$ & $0.61 \pm 0.23$ \\
        NASNetMobile & $0.79 \pm 0.07$ & $0.24 \pm 0.11$ & $0.33 \pm 0.09$ & $\mathbf{0.48 \pm 0.09}$ & $0.42 \pm 0.03$ & $0.48 \pm 0.13$ \\
        ResNet50 & $0.79 \pm 0.05$ & $\mathbf{0.40 \pm 0.16}$ & $0.69 \pm 0.10$ & $\mathbf{0.74 \pm 0.07}$ & $0.54 \pm 0.13$ & $0.80 \pm 0.04$ \\
        ResNet50V2 & $\mathbf{0.85 \pm 0.10}$ & $\mathbf{0.55 \pm 0.13}$ & $0.58 \pm 0.04$ & $\mathbf{0.61 \pm 0.08}$ & $\mathbf{0.74 \pm 0.07}$ & $0.71 \pm 0.02$\\
        VGG16 & $0.74 \pm 0.10$ & $\mathbf{0.62 \pm 0.09}$ & $0.65 \pm 0.06$ & $\mathbf{0.44 \pm 0.18}$ & $0.54 \pm 0.11$ & $0.76 \pm 0.07$\\
        Xception & $\mathbf{0.89 \pm 0.05}$ & $\mathbf{0.52 \pm 0.08}$ & $0.72 \pm 0.06$ & $\mathbf{0.61 \pm 0.11}$ & $\mathbf{0.81 \pm 0.05}$ & $0.82 \pm 0.04$\\ [5pt]
        \textbf{Dermatologists}\\
        Average & $0.95 \pm 0.03$ & $0.67 \pm 0.18$ & $0.88 \pm 0.06$ & $0.59 \pm 0.11$ & $0.88 \pm 0.09$ & $0.92 \pm 0.05$ \\
    \bottomrule
    \end{tabular}
    \label{table:diagnosis_performance_sensitivity}
\end{table}

\begin{table}[h!]
    \setlength{\tabcolsep}{7pt} 
    \renewcommand{\arraystretch}{1.5} 
    \centering
    \caption{Diagnostic performance of ConvNets~(average $\pm$ standard deviation across five runs) and dermatologists~(average $\pm$ standard deviation across eight dermatologists) in terms of specificity on the DermXDB holdout set, split by diagnosis. Several ConvNets achieve expert-level specificity~(in~\textbf{bold}).}
    \begin{tabular}{lrrrrrr}
    \toprule
         & 
         \multicolumn{1}{p{10pt}}{\centering\textbf{Acne}} & 
         \multicolumn{1}{p{10pt}}{\centering \textbf{Actinic} \\ \textbf{keratosis}} & 
         \multicolumn{1}{p{10pt}}{\centering\textbf{Psoriasis}} & 
         \multicolumn{1}{p{10pt}}{\centering \textbf{Seborrheic} \\ \textbf{dermatitis}} & 
         \multicolumn{1}{p{10pt}}{\centering \textbf{Viral} \\ \textbf{warts}}  & 
         \multicolumn{1}{p{10pt}}{\centering \textbf{Vitiligo}} \\
    \midrule
        \textbf{ConvNets}\\
        DenseNet121 & $0.93 \pm 0.01$ & $0.97 \pm 0.01$ & $0.92 \pm 0.01$ & $\mathbf{0.91 \pm 0.02}$ & $0.97 \pm 0.01$ & $0.98 \pm 0.01$ \\
        EfficientNet-B0 & $\mathbf{0.93 \pm 0.06}$ & $0.96 \pm 0.02$ & $0.91 \pm 0.01$ & $\mathbf{0.89 \pm 0.03}$ & $0.95 \pm 0.01$ & $0.95 \pm 0.01$ \\
        InceptionV3 & $0.92 \pm 0.03$ & $0.92 \pm 0.02$ & $\mathbf{0.92 \pm 0.03}$ & $\mathbf{0.91 \pm 0.06}$ & $0.96 \pm 0.02$ & $\mathbf{0.97 \pm 0.03}$\\
        InceptionResNetV2 & $0.94 \pm 0.02$ & $0.97 \pm 0.01$ & $0.86 \pm 0.03$ & $\mathbf{0.86 \pm 0.02}$ & $0.97 \pm 0.01$ & $0.92 \pm 0.04$ \\
        MobileNet & $0.91 \pm 0.05$ & $0.96 \pm 0.02$ & $\mathbf{0.88 \pm 0.07}$ & $\mathbf{0.87 \pm 0.08}$ & $0.97 \pm 0.02$ & $0.94 \pm 0.02$ \\
        MobileNetV2 & $0.66 \pm 0.15$ & $\mathbf{0.99 \pm 0.01}$ & $0.98 \pm 0.03$ & $\mathbf{0.79 \pm 0.11}$ & $\mathbf{1.00 \pm 0.00}$ & $\mathbf{0.94 \pm 0.07}$ \\
        NASNetMobile & $0.65 \pm 0.04$ & $0.97 \pm 0.01$ & $0.96 \pm 0.01$ & $\mathbf{0.86 \pm 0.02}$ & $0.96 \pm 0.03$ & $0.96 \pm 0.03$ \\
        ResNet50 & $0.93 \pm 0.02$ & $0.99 \pm 0.01$ & $0.89 \pm 0.04$ & $\mathbf{0.86 \pm 0.04}$ & $0.97 \pm 0.02$ & $0.96 \pm 0.01$ \\
        ResNet50V2 & $0.90 \pm 0.06$ & $0.95 \pm 0.03$ & $0.92 \pm 0.01$ & $\mathbf{0.90 \pm 0.04}$ & $0.96 \pm 0.02$ & $0.97 \pm 0.01$\\
        VGG16 & $0.90 \pm 0.07$ & $0.92 \pm 0.03$ & $\mathbf{0.90 \pm 0.05}$ & $\mathbf{0.95 \pm 0.04}$ & $0.97 \pm 0.02$ & $0.92 \pm 0.03$\\
        Xception & $0.91 \pm 0.04$ & $0.98 \pm 0.01$ & $\mathbf{0.93 \pm 0.02}$ & $\mathbf{0.92 \pm 0.02}$ & $0.97 \pm 0.02$ & $0.96 \pm 0.03$\\ [5pt]
        \textbf{Dermatologists}\\
        Average & $0.99 \pm 0.01$ & $1.00 \pm 0.00$ & $0.96 \pm 0.02$ & $0.99 \pm 0.01$ & $1.00 \pm 0.00$ & $1.00 \pm 0.00$ \\
    \bottomrule
    \end{tabular}
    \label{table:diagnosis_performance_specificity}
\end{table}

\end{document}